\documentclass{article}

\PassOptionsToPackage{numbers, compress}{natbib}

 \usepackage[main, final]{neurips_2025}

\workshoptitle{NeurIPS 2025 Workshop
on Generative AI in Finance}

\usepackage[utf8]{inputenc} 
\usepackage[T1]{fontenc}    
\usepackage{hyperref}       
\usepackage{url}            
\usepackage{booktabs}       
\usepackage{amsfonts}       
\usepackage{nicefrac}       
\usepackage{microtype}      
\usepackage{xcolor}         
\usepackage{natbib}
\usepackage{longtable}
\title{A Methodology for Assessing the Risk of Metric Failure in LLMs Within the Financial Domain.}

\workshoptitle{GenAI in Finance}
\author{
\textbf{William Flanagan}$^{1}$\thanks{Correspondence to: William Flanagan <william.flanagan@bny.com>}, 
\textbf{Mukunda Das}$^{2}$, 
\textbf{Rajitha Ramanyake}$^{2}$, 
\textbf{Swanuja Maslekar}$^{1}$, \\
\textbf{Meghana Mangipudi}$^{2}$, 
\textbf{Jeel Shah}$^{2}$, 
\textbf{Joong Ho Choi}$^{2}$,
\textbf{Shruti Nair}$^{2}$, \\
\textbf{Shambhavi Bhusan}$^{1}$, 
\textbf{Sanjana Dulam}$^{2}$, 
\textbf{Mouni Pendharkar}$^{1}$,\\ 
\textbf{Nidhi Singh}$^{2}$, 
\textbf{Vashisth Doshi}$^{3}$, 
\textbf{Sachi Shah Paresh}$^{3}$ \\
\\
$^{1}$BNY Responsible AI Office \quad
$^{2}$BNY AI Hub \quad
$^{3}$Carnegie Mellon University
}

\begin{document}

\maketitle

\begin{abstract}
As Generative Artificial Intelligence is adopted across the financial services industry, a
significant barrier to adoption and usage is measuring model performance. Historical
machine learning metrics can oftentimes fail to generalize to GenAI workloads and are
often supplemented using Subject Matter Expert (SME) Evaluation. Even in this combination,
many projects fail to account for various unique risks present in choosing specific
metrics. Additionally, many widespread benchmarks created by foundational research
labs and educational institutions fail to generalize to industrial use. This paper explains
these challenges and provides a Risk Assessment Framework to allow for better
application of SME and machine learning Metrics.
\end{abstract}

\section{Introduction}
Generative AI is quickly becoming one of the most prolific rollouts of advanced technology in banking. Organizations such as BNY, J.P. Morgan, Goldman Sachs, and many more are putting this technology into the hands of thousands of employees~\cite{Bousquette2025DigitalWorkers,Bousquette2025BNY,BNYMellon2025AI}. Many of these employees, however, may have limited technical knowledge. This barrier is overcome through radically usable systems, where technological experience is not as much of a prerequisite to adoption and usage of the technology. While the tools can be adopted by the masses, the validation of individual instantiations of the tools, specifically agentic use cases, is not as accessible. One key barrier is metric explainability, specifically broken into two sub risks: metric individualism and lack of communicability. This barrier is common because academia and research labs often investigate these machine learning (ML) and artificial intelligence (AI) metrics in silos rather than taking input from industry leaders into consideration. Without relevant context, the usage of metrics that attempt to evaluate performance, risk, fairness and bias, etc. can become ineffective or misleading. This in turn leads to situations such as:
\begin{itemize}
    \item Algorithmic bias in credit underwriting (e.g., the Apple Card investigation): High-profile consumers reported that men were offered significantly higher credit limits than their wives, despite shared assets and, in some cases, the wife having a superior credit score~\cite{NYSDFS2021AppleCard}. Although no evidence of unlawful discrimination was found under the Equal Credit Opportunity Act (``ECOA'')~\cite{US1974ECOA}, the Bank's inability to provide transparent reasoning behind the decision at first, along with poor customer service, eroded customer trust~\cite{NYSDFS2021AppleCard}.
    \item Flawed automation in insurance claim adjudication (UnitedHealth \& Cigna Lawsuits): These recent lawsuits assert claims based on, among other things, the use of AI models to deny medical claims without proper human review (i.e., minimal oversight, sometimes as little as 1.2 seconds per claim)~\cite{AMBest2025UnitedHealth,Hansard2023Cigna,HavenHealth2024UnitedHealth}. Here, it is possible that the AI performance was optimized for speed or cost reduction but failed to consider ethical or clinical appropriateness dimensions. It may not have been evaluated in its complete socio-technical context.
\end{itemize}

This lack of clarity alone has unique risks, including a loss of regulatory and employee faith in the robustness of a company's AI ecosystem. It is vital that industry and academia partner to create domain specific metrics that aid organizations in inferring in what ways their models are successful or fail to meet thresholds.

\section{Trust (Regulatory \& Employee)}

Flawed metrics harm the legitimacy of AI use case in production and add risk to firms. One issue, however, is that of a breach in the trust of employees towards the organization. Artificial intelligence is an intimidating topic for many; this can only be overcome through transparent adoption, and thorough communication of the AI systems in question. Working in a financial institution means the trust of clients, consumers and regulators is critical. This trust is dependent on operational precision, such as processing payments of the correct amount, executing trades as instructed, and credit decisions that are explainable. Employees are eager to leverage agentic capabilities with the rapid progress of AI to automate labor intensive tasks. Benchmarks designed to test a large language model's (LLM)  capabilities show great promise. However, an LLM's confident but false responses, hallucinations and lack of contextual awareness can instantly weaken an employee's trust in the system~\cite{Giovine2024Explainability}. Attestations, disclaimers and reminders that ``the model may be wrong'' forces employees to review every output, prompting the question of whether these tools are adding value beyond the traditional process. A further limitation is that LLMs still struggle with natively understanding structured and semi-structured tabular data formats (like spreadsheets, Microsoft Excel files, etc.) -- a format foundational to the industry.

This can lead to two things -- a slower adoption of AI amongst employees or adoption staying the same with an increased occurrence of incorrect responses. In this high-stakes and heavily regulated industry, even one incorrect response can undermine confidence entirely. An employee would be expected  to manually verify every subsequent output, nullifying the efficiency gains in the automation promises~\cite{Giovine2024Explainability}. In other scenarios, incorrect responses can lead to regulatory fines, notices and a lack of trust amongst clients and consumers.

Governance bodies within banks can mandate human oversight for the above reasons. Regulators look for explainable model behavior and the ability to audit it. Closed-source LLMs already limit banks' visibility into its training data and weights, which can hamper model validation. In addition to traditional metrics such as accuracy, precision, F1 scores, ROUGE, BLEU scores, among others, regulators expect banks to develop new metrics specific to the use case as needed, such as hallucination rates and factual consistency.  The EU AI Act will require disclosure of model capabilities, limitations and decisions for high-risk AI systems~\cite{AIAct2024Summary} and Office of the Comptroller of the Currency (OCC) manuals emphasize documenting decision processes~\cite{OCC2025Semiannual}. The black-box nature of LLMs undermines confidence, therefore using explainable AI, human-review processes and clear documentation to see how an LLM arrived at an answer can reduce regulatory pushback and distrust~\cite{IBM2024GenAICompliance}. Due to this, teams need to constantly design tests that can satisfy these requirements to get a green signal from model validation teams - which can be tedious and cumbersome, leading to delays in production of AI solutions. Having industry specific benchmarks tailored to commonly occurring tasks with financial context can eliminate the need to constantly prove AI's capabilities.

\section{Towards Better Generalizable Metrics}
While academic and research organizations such as OpenAI, Anthropic, and DeepMind advance the state of the art, their evaluation methods remain optimized for benchmark performance and scientific milestones rather than production realities. Metrics like MMLU, SWE-bench, or curated reasoning tasks highlight progress toward general intelligence but fail to capture the messiness of real enterprise workloads—inputs with regulatory nuance, shifting data distributions, cost constraints, and latency SLAs. As a result, each new model release may deliver impressive benchmark gains but does not always translate into proportional improvements for industry users. The gap is not that these models are useless, but that lab-level evaluations emphasize ``can the model solve this test?” whereas enterprises must ask ``can the system deliver reliable, auditable, cost-effective outputs under real conditions?''. This often forces organizations to rely on costly evaluations from SMEs. However, the SME evaluation can introduce subjectivity. To overcome these challenges, it is essential to combine both with adaptive methodologies that can evolve alongside changing conditions and help prevent evaluation failures after deployment. The following methods can be incorporated into the evaluation system to support more informed decision-making:

\begin{itemize}
    \item AI models can degrade over time due to shifting market conditions or evolving human behaviors. To address this, automated continuous monitoring should be employed, using statistical tests to detect both concept and data drift. When anomalies are detected, alerts can be triggered for human review and intervention.
    \item Traditional backtesting is limited to historical conditions, making models vulnerable to unforeseen events. To improve robustness, adversarial and catastrophic stress testing should be conducted prior to deployment. Synthetic data generation can simulate severe but plausible scenarios (e.g., market crashes, data drift), helping to identify breaking points before they lead to high-risk failures in the real world.
    \item A dedicated agent should be deployed to detect out-of-distribution operations. This agent would flag novel or unprecedented circumstances, ensuring that humans can intervene when necessary. Such a safeguard reduces the risk of models making overconfident yet incorrect predictions.
\end{itemize}

These measures not only enhance model reliability but also foster greater customer trust.

Bridging this gap requires evaluation frameworks that are explicitly use-case centric. Benchmark scores alone are insufficient; financial institutions need metrics that capture task success, compliance fidelity, error severity, and operational viability.
This means combining scalable techniques like ``LLM-as-Judge'' for broad coverage with SME sampling for depth as proposed in the Global AI Assurance Pilot~\cite{Turing2025GenerativeTrustworthy}, while layering in business-aligned KPIs such as cost per accepted resolution, critical error rates, and escalation recall. The key is not to discard lab metrics, but to contextualize them within enterprise-specific evaluation stacks that make results communicable to compliance officers, executives, and frontline staff alike. Only when academic progress is continuously re-grounded in business KPIs and regulatory requirements will the optimism of research translate into durable enterprise value.

\section{Risk Classification}
To proactively identify, assess, and remediate the ways in which evaluation metrics themselves can fail -- undermining model validation, regulatory compliance, and end-user trust, this classification (Table~\ref{tab:LLM-risk-classes-full}) organizes common ``metric failure” modes into five high-level categories. This classification can be leveraged to better understand how metrics fail, and aid institutions in making more informed decisions. 

\section{Limitations}
This work is limited in scope to generative AI applications within finance. Although many
risks proposed may generalize across domains, the severity of such risks may differ
depending upon the industry in which a model is being leveraged. This is also true even
within finance. Determinations of risk level and probability were additionally made by
SMEs within the space, based off the use cases pursued within our
organization.

\begin{longtable}{p{1.2cm} p{1.7cm} p{3.2cm} p{1.2cm} p{0.9cm} p{3.2cm}}
\toprule
Category & Failure Mode & Definition & Probability & Impact & Mitigation \\
\midrule
Data Risk &
Distribution Shift &
Inputs drift over time away from the data slice used to calibrate the metric &
High &
High &
Set up automated data-drift detectors and periodic metric re-validation \\

Data Risk &
Label Drift &
SME judgments evolve (e.g., new guidelines change what ``factual'' means) &
Medium &
Medium &
Maintain versioned annotation guidelines and track inter-annotator agreement \\

Model Risk &
Calibration Drift &
Score distributions shift between model versions, masking real performance degradation &
Medium &
High &
Deploy control charts; trigger automated re-calibration when distributions exceed thresholds \\

 &
Adversarial Vulnerability &
Small input perturbations cause large deviations in metric outputs &
Low &
High &
Harden pre-processing; fuzz-test metrics with adversarial examples \\

Process \& Annotation Risk &
Annotation Inconsistency  &
SMEs apply rubric unevenly across samples or over time &
Medium &
High &
Enforce detailed rubrics, cross-review sampling, and compute inter-annotator agreement metrics \\

&
Action Bias &
SMEs knowingly or unknowingly over-score to accelerate model approval &
Medium &
High &
Introduce blinded review, rotate reviewers, and audit for bias patterns \\

&
Scope Misalignment &
SMEs evaluate based on personal expectations rather than the model's intended capabilities &
Medium &
Medium &
Provide clear task descriptions and run calibration workshops before scoring\\

&
Scalability Constraints &
Limited SME bandwidth forces sampling of only a handful of outputs, missing rare but critical failure cases&
High&
Medium&
Blend automated checks (“LLM-as-judge”) with periodic SME deep dives\\

Governance \& Compliance Risk&
Documentation Gaps&
Metric definitions, version history, and ownership are incomplete or not centrally stored&
High&
High&
Use a centralized metric registry, enforce changelog entries, assign a metric steward\\

&
Knowledge Continuity Risk&
SME turnover leaves team without historical context for metric rationale&
High&
High&
Embed metric rationale in documentation; conduct regular knowledge-transfer sessions\\

&
Domain-Intensive Metrics&
Metrics use domain jargon and complex formulas that non-SMEs cannot interpret&
Medium&
High&
Partner with compliance to co-author metric specifications and publish explainable guides\\

&
Regulatory Misalignment&
Metrics fail to capture or adapt to evolving regulatory requirements&
Medium&
High&
Maintain a regulatory watchlist; update evaluation criteria as rules evolve\\

Ethical \& Reputational Risk&
Bias \& Fairness Failures&
Metrics systematically under-score outputs for certain demographic or business segments&
Medium&
High&
Audit metrics for disparate impact; incorporate fairness constraints into scoring procedures\\

&
Hallucination Escape&
Metrics do not penalize confident but factually incorrect assertions&
High&
High&
Add factuality checks, counterfactual tests, and human-review triggers for high-risk outputs\\

\bottomrule
\caption{Risk Classification}
\label{tab:LLM-risk-classes-full}
\end{longtable}

\bibliographystyle{plainnat}
\bibliography{references}

\begin{thebibliography}{13}
\providecommand{\natexlab}[1]{#1}
\providecommand{\url}[1]{\texttt{#1}}
\expandafter\ifx\csname urlstyle\endcsname\relax
  \providecommand{\doi}[1]{doi: #1}\else
  \providecommand{\doi}{doi: \begingroup \urlstyle{rm}\Url}\fi

\bibitem[{AM Best Company}(2025)]{AMBest2025UnitedHealth}
{AM Best Company}.
\newblock Class action alleges unitedhealth used ai to wrongly deny medicare advantage claims.
\newblock Best's News, February 18, 2025, 2025.
\newblock URL \url{https://news.ambest.com/newscontent.aspx?AltSrc=104\&RefNum=264200}.

\bibitem[{Bank of New York Mellon}(2025)]{BNYMellon2025AI}
{Bank of New York Mellon}.
\newblock Artificial intelligence.
\newblock Bank of New York Mellon Corporate Website, 2025.
\newblock URL \url{https://www.bny.com/corporate/global/en/about-us/technology-innovation/artificial-intelligence.html}.

\bibitem[Bousquette(2025{\natexlab{a}})]{Bousquette2025BNY}
Isabella Bousquette.
\newblock {BNY, America's Oldest Bank, Signs Multiyear Deal With OpenAI}.
\newblock The Wall Street Journal, February 26, 2025, 2025{\natexlab{a}}.
\newblock URL \url{https://www.wsj.com/articles/bny-americas-oldest-bank-signs-multiyear-deal-with-openai-74987d1d}.

\bibitem[Bousquette(2025{\natexlab{b}})]{Bousquette2025DigitalWorkers}
Isabella Bousquette.
\newblock {'Digital Workers' Have Arrived in Banking}.
\newblock The Wall Street Journal, June 30, 2025, 2025{\natexlab{b}}.
\newblock URL \url{https://www.wsj.com/articles/digital-workers-have-arrived-in-banking-bf62be49}.

\bibitem[{European Commission}(2024)]{AIAct2024Summary}
{European Commission}.
\newblock High-level summary of the ai act.
\newblock EU Artificial Intelligence Act Website, February 27, 2024 (updated May 30, 2024), 2024.
\newblock URL \url{https://artificialintelligenceact.eu/high-level-summary/}.

\bibitem[Giovine et~al.(2024)Giovine, Roberts, Pometti, and Bankhwal]{Giovine2024Explainability}
Carlo Giovine, Roger Roberts, Mara Pometti, and Medha Bankhwal.
\newblock Building ai trust: The key role of explainability.
\newblock McKinsey \& Company Insights, November 26, 2024, 2024.
\newblock URL \url{https://www.mckinsey.com/capabilities/quantumblack/our-insights/building-ai-trust-the-key-role-of-explainability}.

\bibitem[Hansard(2023)]{Hansard2023Cigna}
Sara Hansard.
\newblock Cigna sued over alleged automated patient claims denials.
\newblock Bloomberg Law, July 24, 2023, 2023.
\newblock URL \url{https://news.bloomberglaw.com/health-law-and-business/cigna-sued-over-alleged-automated-patient-claims-denials}.

\bibitem[{Haven Health Management}(2024)]{HavenHealth2024UnitedHealth}
{Haven Health Management}.
\newblock Unitedhealth lawsuit: Can ai deny healthcare services?
\newblock Haven Health Management Blog, December 06, 2024, 2024.
\newblock URL \url{https://havenhealthmgmt.org/unitedhealth-lawsuit-can-ai-deny-healthcare-services/}.

\bibitem[{IBM Corporation}(2024)]{IBM2024GenAICompliance}
{IBM Corporation}.
\newblock Maximizing compliance: Integrating gen ai into the financial regulatory framework.
\newblock IBM THINK Insights, August 12, 2024, 2024.
\newblock URL \url{https://www.ibm.com/think/insights/maximizing-compliance-integrating-gen-ai-into-the-financial-regulatory-framework}.

\bibitem[{New York State Department of Financial Services}(2021)]{NYSDFS2021AppleCard}
{New York State Department of Financial Services}.
\newblock Report on apple card investigation.
\newblock NYS Department of Financial Services Report, March 2021.
\newblock URL \url{https://www.dfs.ny.gov/reports_and_publications/202103_report_apple_card_investigation}.

\bibitem[{Office of the Comptroller of the Currency (U.S. Department of the Treasury)}(2025)]{OCC2025Semiannual}
{Office of the Comptroller of the Currency (U.S. Department of the Treasury)}.
\newblock Semiannual risk perspective, spring 2025.
\newblock Office of the Comptroller of the Currency Report, May 2025, 2025.
\newblock URL \url{https://www.occ.gov/publications-and-resources/publications/semiannual-risk-perspective/files/pub-semiannual-risk-perspective-spring-2025.pdf}.

\bibitem[{The Alan Turing Institute}(2025)]{Turing2025GenerativeTrustworthy}
{The Alan Turing Institute}.
\newblock Making generative ai trustworthy and reliable for adoption at scale.
\newblock Turing Institute Blog, June 16, 2025, 2025.
\newblock URL \url{https://www.turing.ac.uk/blog/making-generative-ai-trustworthy-and-reliable-adoption-scale}.

\bibitem[{U.S. Congress}(1974)]{US1974ECOA}
{U.S. Congress}.
\newblock {Equal Credit Opportunity Act (ECOA)}.
\newblock 15 U.S.C. § 1691, 1974.
\newblock URL \url{https://tinyurl.com/USECOA}.
\newblock [Online; accessed 28-Aug-2025].

\end{thebibliography}


\newpage

\end{document}